\documentclass[10pt,twocolumn,letterpaper]{article}

\usepackage{cvpr}              

\usepackage{multirow}
\usepackage{colortbl,booktabs}
\usepackage{makecell}

\usepackage[dvipsnames]{xcolor}

\definecolor{cvprblue}{rgb}{0.21,0.49,0.74}

\usepackage[pagebackref,breaklinks,colorlinks,citecolor=cvprblue]{hyperref}

\def\paperID{13178} 
\def\confName{CVPR}
\def\confYear{2024}

\title{Rethinking Inductive Biases for Surface Normal Estimation}

\author{Gwangbin Bae \;\;\;\; Andrew J. Davison\\
Dyson Robotics Lab, Imperial College London\\
{\tt\small \{g.bae, a.davison\}@imperial.ac.uk}
}

\definecolor{myyellow}{rgb}{0.8,0.8,0}
\definecolor{mygreen}{rgb}{0,0.8,0}
\definecolor{myred}{rgb}{0.8,0,0}

\begin{document}

\twocolumn[{%
\renewcommand\twocolumn[1][]{#1}%
\maketitle
\includegraphics[width=1.0\textwidth]{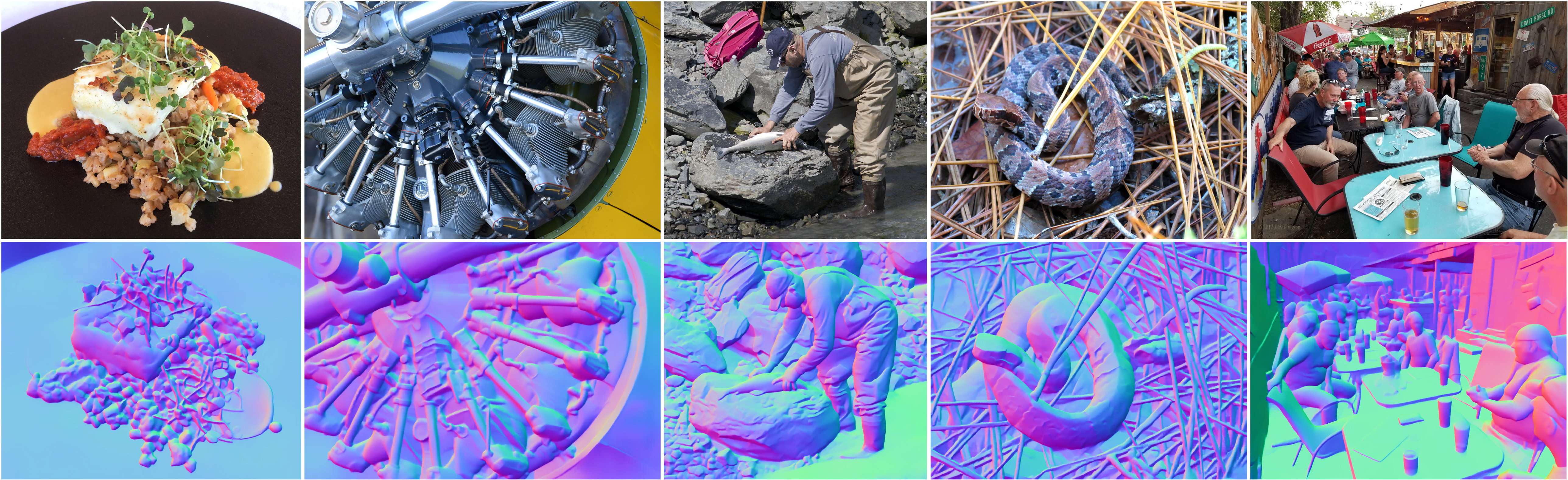}
\captionof{figure}{\textbf{Examples of challenging in-the-wild images and their surface normals predicted by our method.} \vspace{2em}}
\label{fig:teaser}
}]

\begin{abstract}
Despite the growing demand for accurate surface normal estimation models, existing methods use general-purpose dense prediction models, adopting the same inductive biases as other tasks. In this paper, we discuss the inductive biases needed for surface normal estimation and propose to (1) utilize the per-pixel ray direction and (2) encode the relationship between neighboring surface normals by learning their relative rotation. The proposed method can generate crisp --- yet, piecewise smooth --- predictions for challenging in-the-wild images of arbitrary resolution and aspect ratio. Compared to a recent ViT-based state-of-the-art model, our method shows a stronger generalization ability, despite being trained on an orders of magnitude smaller dataset. The code is available at \url{https://github.com/baegwangbin/DSINE}.
\end{abstract}    
\section{Introduction}
\label{sec:intro}

We address the problem of estimating per-pixel surface normal from a single RGB image. This task, unlike monocular depth estimation, is not affected by scale ambiguity and has a compact output space (a unit sphere vs. positive real value), making it feasible to collect data that densely covers the output space. As a result, learning-based surface normal estimation methods show strong generalization capability for out-of-distribution images, despite being trained on relatively small datasets~\cite{SNfromRGB_2021_EESNU}.

Despite their essentially local property, predicted surface normals contain rich information about scene geometry. In recent years, their usefulness has been demonstrated for various computer vision tasks, including image generation~\cite{2023_ControlNet}, object grasping~\cite{2023_MonoGraspNet}, multi-task learning~\cite{2023_Prismer}, depth estimation~\cite{SNfromRGB_2018_GeoNet, 2022_IronDepth}, simultaneous localization and mapping~\cite{2023_NICER-SLAM}, human body shape estimation~\cite{2022_ICON, 2023_ECON,2023_FOUND}, and CAD model alignment~\cite{2022_SPARC}. However, despite the growing demand for accurate surface normal estimation models, there has been little discussion on the right inductive biases needed for the task.

State-of-the-art surface normal estimation methods~\cite{SNfromRGB_2021_EESNU, MDE_2021_TransDepth, 2021_Omnidata, 2022_3DCC} use general-purpose dense prediction models, adopting the same inductive biases as other tasks (e.g. depth estimation and semantic segmentation). For example, CNN-based models~\cite{SNfromRGB_2021_EESNU, 2021_Omnidata} assume translation equivariance and use the same set of weights for different parts of the image. While such weight-sharing can improve sample efficiency~\cite{1998_LeCun_CNN}, it is sub-optimal for surface normal estimation as a pixel's \textit{ray direction} provides important cues and constraints for its surface normal. This has limited the accuracy of the prediction and the ability to generalize to images taken with out-of-distribution cameras.

Another important aspect of surface normal estimation overlooked by existing methods is that there are common typical relationships between the normals at nearby image pixels. It is well understood that many 3D objects in a scene are piece-wise smooth~\cite{1994_Hoppe_piecewise_smooth} and that neighboring normals often have similar values. There is also a very frequently occurring relationship between groups of nearby pixels on two surfaces in contact, or between groups of pixels on a continuously curving surface: their normals are related by a rotation through a certain angle, about an axis lying within the surface and which is sometimes visible in the image as an edge. 


In this paper, we provide a thorough discussion of the inductive biases needed for deep learning-based surface normal estimation and propose three architectural changes to incorporate such biases:

\begin{itemize}
    \item We supply \textit{dense pixel-wise ray direction} as input to the network to enable camera intrinsics-aware inference and hence improve the generalization ability. 
    \item We propose a \textit{ray direction-based activation function} to ensure the visibility of the prediction.
    \item We recast surface normal estimation as \textit{rotation estimation}, where the relative rotation with respect to the neighboring pixels is estimated in the form of axis-angle representation. This allows the model to generate predictions that are piece-wise smooth, yet crisp at the intersection between surfaces.
\end{itemize}

The proposed method shows strong generalization ability. It can generate highly detailed predictions even for challenging in-the-wild images of arbitrary resolution and aspect ratio (see Fig.~\ref{fig:teaser}). We outperform a recent ViT-based state-of-the-art method~\cite{2021_Omnidata,2022_3DCC} --- both quantitatively and qualitatively --- despite being trained on an orders of magnitude smaller dataset. 

\section{Related work}
\label{sec:related_work}

Hoiem et al.~\cite{SNfromRGB_2005_Pop_up, SNfromRGB_2007_Hoiem} were among the first to propose a learning-based approach for monocular surface normal estimation. The output space was discretized and handcrafted features were extracted to classify the normals. Fouhey et al.~\cite{SNfromRGB_2013_3DP} took a different approach and tried to detect geometrically informative \textit{primitives} from data. For detected primitives, the normal maps of the corresponding training patches were aligned to recover a dense prediction. Another common approach was to assume a Manhattan World~\cite{MWVP_2000_MW} to adjust the initial prediction~\cite{SNfromRGB_2013_3DP} or generate candidate normals from pairs of vanishing points~\cite{SNfromRGB_2014_Fouhey}.

Following the success of deep convolutional neural networks in image classification~\cite{OTHER_2012_AlexNet}, many deep learning-based methods~\cite{MDE_2015_Eigen, SNfromRGB_2015_Deep3D, SNfromRGB_2016_SkipNet} were introduced. Since then, notable contributions have been made by exploiting the surface normals computed from Manhattan lines~\cite{SNfromRGB_2020_VPLNet}, introducing a spatial rectifier to handle tilted images~\cite{SNfromRGB_2020_TiltedSN}, and estimating the aleatoric uncertainty to improve the performance on small structures and near object boundaries~\cite{SNfromRGB_2021_EESNU}.

Eftekhar et al.~\cite{2021_Omnidata} trained a U-Net~\cite{SEG_2015_UNET} on more than 12 million images covering diverse scenes and camera intrinsics. They recently released an updated model by training a transformer-based model~\cite{MDE_2021_DPT} with sophisticated 3D data augmentation~\cite{2022_3DCC} and cross-task consistency~\cite{2020_XTC}. This model is the current state-of-the-art in surface normal estimation and will be the main comparison for our method.

\section{Inductive bias for surface normal estimation}
\label{sec:inductive_bias}

\begin{figure*}[t!]
\centering
\includegraphics[width=1.0\linewidth]{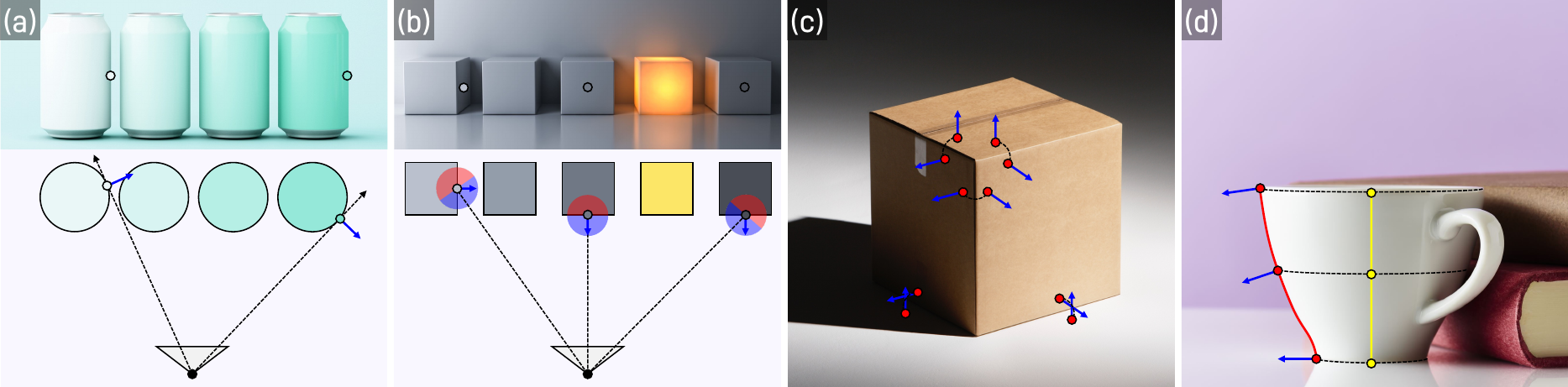}
\caption{\textbf{Motivation.} In this paper, we propose to utilize the \textit{per-pixel ray direction} and estimate the surface normals by learning the \textit{relative rotation between nearby pixels}. \textbf{(a)} Ray direction serves as a useful cue for pixels near occluding boundaries as the normal should be perpendicular to the ray. \textbf{(b)} It also gives us the range of normals that would be visible, effectively halving the output space. \textbf{(c)} The surface normals of certain scene elements --- in this case, the floor --- may be difficult to estimate due to the lack of visual cues. Nonetheless, we can infer their normals by learning the pairwise relationship between nearby normals (e.g. which surfaces should be perpendicular). \textbf{(d)} Modeling the relative change in surface normals is not just useful for flat surfaces. In this example, the relative angle between the normals of the yellow pixels can be inferred from that of the red pixels assuming circular symmetry.}
\label{fig:motivation}
\end{figure*}

In this section, we discuss the inductive biases needed for surface normal estimation. Throughout the rest of this paper, we use the right-hand convention for camera-centered coordinates, where the $X$, $Y$, and $Z$ axes point right, down, and front, respectively.

\subsection{Encoding per-pixel ray direction}

Under perspective projection, each pixel is associated with a ray that passes through the camera center and intersects the image plane at the pixel. Assuming a pinhole camera, a ray of unit depth for a pixel at $(u,v)$ can be written as

\begin{equation}
\label{eqn:ray}
\mathbf{r}(u,v) = \begin{bmatrix}
\frac{u-c_u}{f_u} & \frac{v-c_v}{f_v} & 1
\end{bmatrix}^\intercal,
\end{equation}

\noindent
where $f_u$ and $f_v$ are the focal lengths and $(c_u, c_v)$ are the pixel coordinates of the principal point.

Per-pixel ray direction is essential for surface normal estimation. For rectangular structures (e.g. buildings), we can identify sets of parallel lines and their respective vanishing points. The ray direction at the vanishing point then gives us the 3D orientation of the lines and hence the surface normals~\cite{2003_BOOK_MVG}. Early works on single-image 3D reconstruction \cite{1997_TIP, 2005_Kosecka_rectangular_structures, 2009_Geometric_reasoning, SNfromRGB_2014_Fouhey} made explicit use of such cues.

Now consider an occluding boundary created by a smooth (i.e. infinitely differentiable) surface. As Marr~\cite{1977_Marr_contour} pointed out, the surface normals at an occluding boundary can be determined uniquely by forming a generalized cone (whose apex is at the camera center) that intersects the image plane at the boundary. In other words, the normals at the boundary should be perpendicular to the ray direction (see Fig.~\ref{fig:motivation}-a). Such insights have been widely adopted for under-constrained 3D reconstruction tasks such as single-image shape from shading~\cite{1981_Numerical_SfS}. 

Lastly, the ray direction decides the range of normals that would be \textit{visible} in that pixel, effectively halving the output space (see Fig.~\ref{fig:motivation}-b). This is analogous to the case of depth estimation, where the output should be positive. Such an inductive bias is often adopted by interpreting the network output as log depth~\cite{MDE_2014_Eigen} or by using a ReLU activation~\cite{2018_ReLU_for_MDE}. 


Despite the aforementioned usefulness, state-of-the-art methods~\cite{SNfromRGB_2021_EESNU, 2021_Omnidata} do not encode the ray direction and use CNNs with translational weight sharing, preventing the model from learning ray direction-aware inference. While recent transformer-based models~\cite{MDE_2021_TransDepth, 2022_3DCC} have the capability of encoding the ray direction in the form of learned positional embedding, it is not trivial to inter/extrapolate the positional embedding when testing the model on images taken with out-of-distribution intrinsics.

\subsection{Modeling inter-pixel constraints}

Consider a pixel $i$ and its neighboring pixel $j$. Since their surface normals have unit length and share the same origin (camera center), they are related by a 3D rotation matrix $R \in SO(3)$. While there are different ways to parameterize $R$, we choose the \textit{axis-angle} representation, $\boldsymbol{\theta} = \theta \mathbf{e}$, where a unit vector $\mathbf{e}$ represents the axis of rotation and $\theta$ is the angle of rotation. Then, the exponential map $\exp: \mathfrak{so}(3) \rightarrow SO(3)$ --- which is readily available in modern deep learning libraries~\cite{2015_tensorflow,2020_pytorch3d} --- can map $\boldsymbol{\theta}$ back to $R$.

Within flat surfaces (which are prevalent in man-made scenes/objects), $\theta$ would be zero and $R$ would simply be the identity. In a typical indoor scene, the surfaces of objects are often perpendicular or parallel to the ground plane, creating lines across which the normals should rotate by $90^\circ$ (see Fig.~\ref{fig:motivation}-c). For a curved surface, the relative angle between the pixels can be inferred from the occluding boundaries by assuming a certain level of symmetry (see Fig.~\ref{fig:motivation}-d).

But why should we learn $R$ instead of directly estimating the normals? Firstly, learning the relative rotation is much easier, as the angle between the normals, unlike the normals themselves, is independent of the viewing direction (it is also zero --- or close to zero --- for most pixel pairs). Finding the axis of rotation is also straightforward. When two (locally) flat surfaces intersect at a line, the normals rotate around that intersection. As the image intensity generally changes sharply near such intersections, the task can be as simple as edge detection.

Secondly, the estimated rotation can help improve the accuracy for surfaces with limited visual cues. For instance, while it is difficult to estimate the normal of a texture-less surface, the objects that are in contact with the surface can provide evidence for its normal (see Fig.~\ref{fig:motivation}-c).

Lastly, as long as the relative rotations between the normals are captured correctly, any misalignment between the prediction and the ground truth can be resolved via a single global rotation. For example, in the case of a flat surface, estimating inaccurate but constant normals is better than estimating accurate but noisy normals, as the orientation can easily be corrected (e.g. via sparse depth measurements or visual odometry). This is again analogous to depth estimation where a relative depth map is easier to learn and can be aligned via a global scaling factor.

\section{Our approach}
\label{sec:approach}

From Sec.~\ref{sec:approach-ray} to \ref{sec:approach-nrn}, we explain how a dense prediction network can be modified to encode the inductive biases discussed in Sec.~\ref{sec:inductive_bias}. We then explain the network architecture and our training dataset in Sec.~\ref{sec:approach-arch} and \ref{sec:approach-dataset}.


\subsection{Ray direction encoding}
\label{sec:approach-ray}

\begin{figure}[t]
\centering
\includegraphics[width=1.0\linewidth]{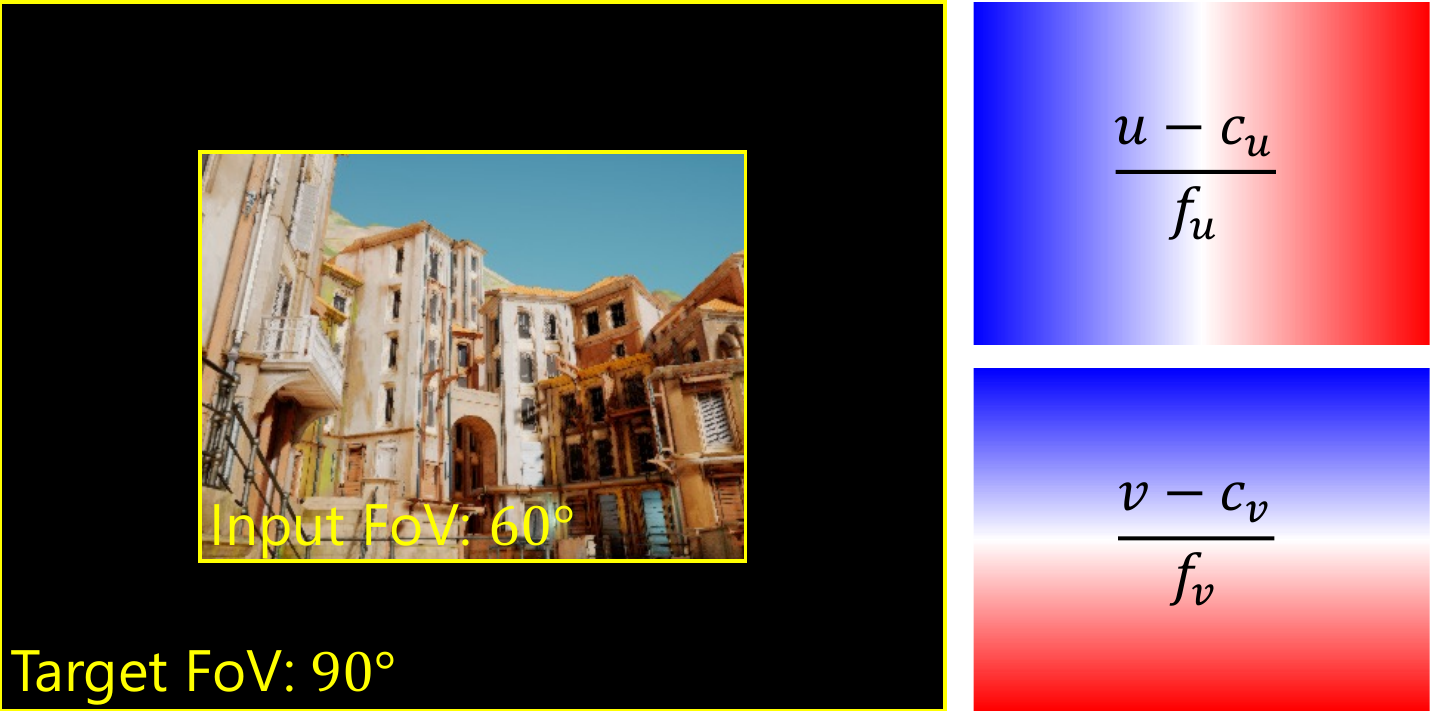}
\caption{\textbf{Encoding camera intrinsics.} \textbf{(left)} To avoid having to learn camera intrinsics-aware prediction, one can zero-pad or crop the images such that they always have the same intrinsics. \textbf{(right)} Instead, we compute the focal length-normalized image coordinates and provide them as additional input to the network.}
\label{fig:method-intrins}
\end{figure}

To avoid having to learn ray direction-aware prediction, one can crop and zero-pad the images such that the principal point is at the center and the field of view is always $\theta^\circ$, where $\theta$ is set to some high value. Then, a pixel at $(u,v)$ will always have the same ray direction, allowing the network to encode it, e.g., in the form of position embedding. However, such an approach (1) wastes the compute for the zero-padded regions, (2) loses high-frequency details from downsampling, and (3) cannot be applied to images with wider field-of-view. Instead, we compute the focal length-normalized image coordinates --- i.e. Eq.~\ref{eqn:ray} --- and provide this as an additional input to the intermediate layers of the network (see Fig.~\ref{fig:method-intrins}). This encoding is similar to that of CAM-Convs~\cite{2019_CAM-Convs} which was designed for depth estimation. Unlike \cite{2019_CAM-Convs}, we do not encode the image coordinates themselves and only encode the ray direction.


\subsection{Ray direction-based activation}
\label{sec:approach-relu}

A surface that is facing \textit{away} from the camera would simply not be visible in the image. An important constraint for surface normal estimation should thus be that the angle between the ray direction and the estimated normal vector must be greater than $90^\circ$. To incorporate such a bias, we propose a ray direction-based activation function analogous to ReLU. Given the estimated normal $\mathbf{n}$ and ray direction $\mathbf{r}$ (both are normalized), the activation can be written as

\begin{equation}
\label{eqn:relu}
\sigma_\text{ray}(\mathbf{n}, \mathbf{r})
:= 
\frac{\mathbf{n} + \left(
\min(0, \mathbf{n} \cdot \mathbf{r}) 
- \mathbf{n} \cdot \mathbf{r}
\right) \mathbf{r}}
{\lVert
\mathbf{n} + \left(
\min(0, \mathbf{n} \cdot \mathbf{r}) 
- \mathbf{n} \cdot \mathbf{r}
\right) \mathbf{r}
\rVert}.
\end{equation}

\begin{figure}[t]
\centering
\includegraphics[width=1.0\linewidth]{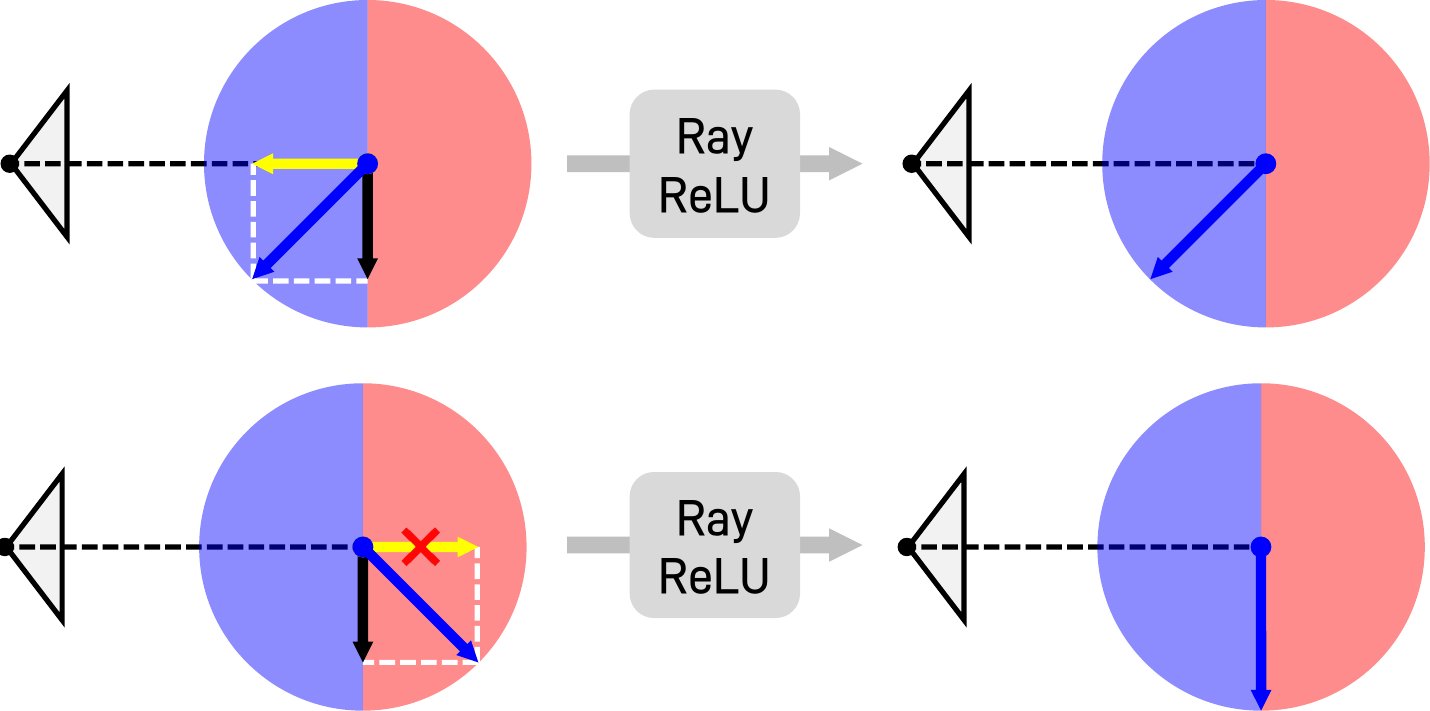}
\caption{\textbf{Ray ReLU activation.} An important constraint for surface normal estimation is that the predicted normal should be \textit{visible}. We achieve this by zeroing out the component that is in the direction of the ray.}
\label{fig:method-relu}
\end{figure}

Eq.~\ref{eqn:relu} ensures that $\mathbf{n} \cdot \mathbf{r} = \cos \theta$ (i.e. the magnitude of $\mathbf{n}$ along $\mathbf{r}$) is less than or equal to zero. The rectified normal is then re-normalized to have a unit length. This is illustrated in Fig.~\ref{fig:method-relu}.

\begin{figure*}[t!]
\centering
\includegraphics[width=1.0\linewidth]{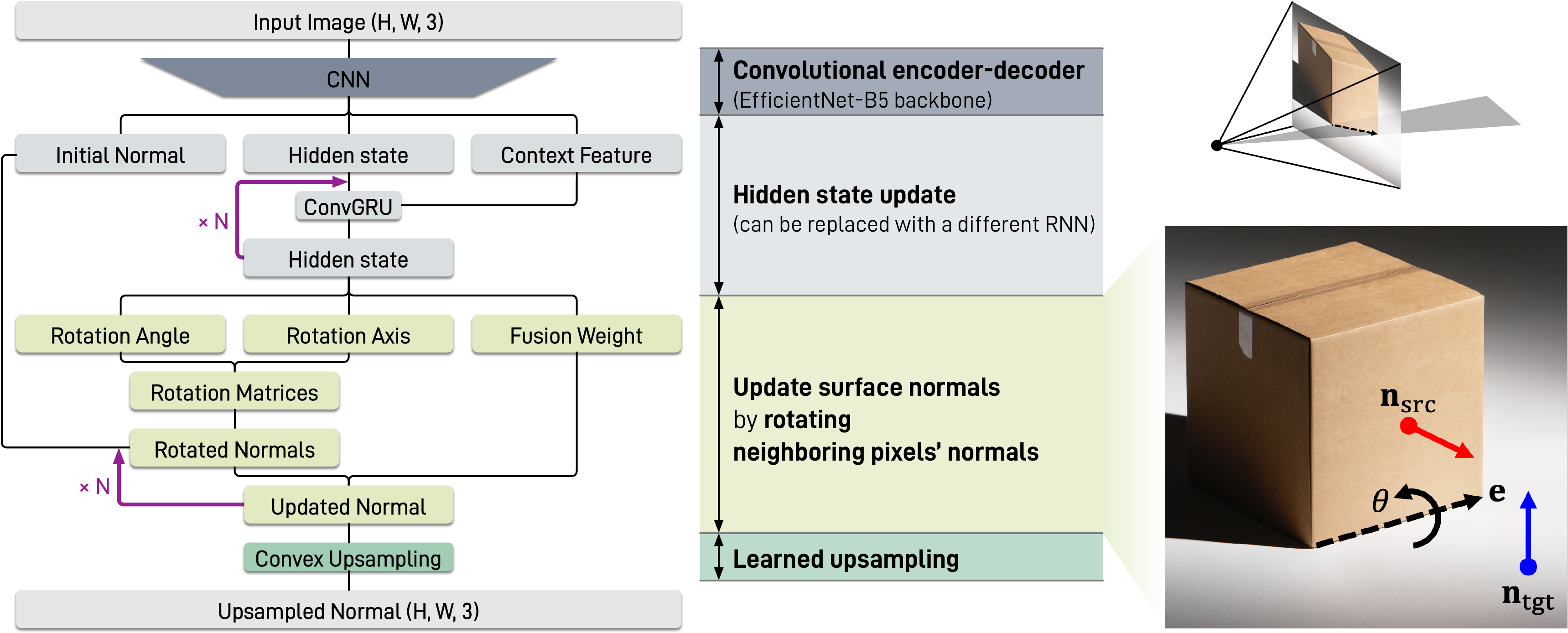}
\caption{\textbf{Network architecture.} A lightweight CNN extracts a low-resolution feature map, from which the initial normal, hidden state and context feature are obtained. The hidden state is then recurrently updated using a ConvGRU~\cite{GRU} unit. From the updated hidden state, we estimate three quantities: rotation angle and axis to define a pairwise rotation matrix for each neighboring pixel; and a set of weights that will be used to fuse the rotated normals.}
\label{fig:method-arch}
\end{figure*}


\subsection{Recasting surface normal estimation as rotation estimation}
\label{sec:approach-nrn}

For pixel $i$, we can define its local neighborhood  $\mathcal{N}_i=\{ j \;:\; |u_i-u_j| \leq \beta \; \text{and} \; |v_i-v_j| \leq \beta \}$. We can then learn the pairwise relationship between the surface normals $\mathbf{n}_i$ and $\mathbf{n}_j$ in the form of a rotation matrix $R_{ij}$.

For each pair of pixels, three quantities should be estimated: First is the angle $\theta_{ij}$ between the two normals. This is easy to learn as $\theta_{ij}$ is independent of the viewing direction and is $0^\circ$ or $90^\circ$ for many pixel pairs. Secondly, we need to estimate the axis of rotation $\mathbf{e}_{ij}$ (i.e. a 3D unit vector around which the normals rotate). While directly learning $\mathbf{e}_{ij}$ requires complicated 3D reasoning, we propose a simpler approach that only requires 2D information.

Let's first consider the case where $\mathbf{n}_i$ and $\mathbf{n}_j$ are on the same smooth surface. As the surface can locally be approximated as a plane, the angle should be close to zero. In such a case, finding the axis is less important as the rotation matrix will be close to the identity. 

Another possibility is that the two points are on different smooth surfaces that are intersecting with each other. In this case, we only need to estimate the \textit{2D projection} of $\mathbf{e}_{ij}$. Suppose that this 2D projection is a vector whose endpoints are $(u_j, v_j)$ and $(u_j + \delta u_{ij}, v_j + \delta v_{ij})$. The 3D vector $\mathbf{e}_{ij}$ should then lie on a plane that passes through the camera center and the two endpoints (see Fig~\ref{fig:method-arch}-right). Formally, this can be written as,

\begin{equation}
\label{eqn:rot_axis}
\begin{aligned}
\mathbf{e}_{ij}
&= X^c(u_j + \delta u_{ij}, v_j + \delta v_{ij}) - X^c(u_j, v_j) \\
&=
\begin{bmatrix}
(z_j+\delta z_{ij}) \cdot \frac{u_j + \delta u_{ij} - c_u}{f_u} 
-
z_j \cdot \frac{u_j - c_u}{f_u} 
\\
(z_j+\delta z_{ij}) \cdot \frac{v_j + \delta v_{ij} - c_v}{f_v} 
-
z_j \cdot \frac{v_j - c_v}{f_v}
\\
\delta z_{ij}
\end{bmatrix}
\end{aligned}
~,
\end{equation}

\noindent
where $X^c(\cdot,\cdot)$ are the camera-centered coordinates corresponding to the pixel and $\delta z$ represents the change in depth. There are two unknowns in Eq.~\ref{eqn:rot_axis}: $z$ and $\delta z$. The first constraint for solving Eq.~\ref{eqn:rot_axis} is that $\lVert \mathbf{e}_{ij} \rVert = 1$. We can then assume that the surface normal $\mathbf{n}_j$ is perpendicular to $\mathbf{e}_{ij}$ (i.e. $\mathbf{n}_j \cdot \mathbf{e}_{ij} = 0$), which is true for pixels near the intersection between two (locally) flat surfaces. Such an approach is appealing for 2D CNNs --- whose initial layers are known to be oriented edge filters --- as the image intensity tends to change sharply near the intersection between two surfaces. 

The final remaining possibility for $\mathbf{n}_i$ and $\mathbf{n}_j$ is that they are on two disconnected surfaces or on non-smooth surfaces. As estimating $R_{ij}$ in such a case is significantly more challenging than the other two scenarios, we choose to down-weight them with a set of weights $\{ w_{ij} \}$, which is the last quantity we learn. The updated normal of pixel $i$ can be written as

\begin{equation}
\label{eqn:rot_final}
\begin{aligned}
\mathbf{n}^{t+1}_i &=
\frac{\sum_j w_{ij} \sigma_\text{ray} (R_{ij} \mathbf{n}^{t}_j, \mathbf{r}_i)}{\lVert \sum_j w_{ij} \sigma_\text{ray}(R_{ij} \mathbf{n}^{t}_j, \mathbf{r}_i) \rVert} \\
R_{ij} &= \exp \left(
\theta_{ij} [\mathbf{e}_{ij}]_\times
\right).
\end{aligned}
\end{equation}

\noindent
where the ray-ReLU activation, introduced in Sec.~\ref{sec:approach-relu}, is used to ensure that the rotated normals are in the visible range for the target pixel $i$. We also added a superscript for the normals to represent an iterative update.

To summarize, given some initial surface normal prediction $\mathbf{n}^t$, the network should estimate the following three quantities --- for each pixel $i$ --- in order to obtain the updated normal map $\mathbf{n}^{t+1}$:

\begin{itemize}
    \item The rotation angles $\{ \theta_{ij}$ \} for the neighboring pixels. For the output, we use a sigmoid activation followed by a multiplication of $\pi$. 
    \item 2D unit vectors $\{ (\delta u_{ij}, \delta v_{ij}) \}$ whose orientation represents the 2D projection of the rotation axes $\{ \mathbf{e}_{ij} \}$. We use two output channels followed by an L2 normalization. This, combined with $\{ \mathbf{n}^t_j \}$ gives us the axes of rotation.
    \item The weights $\{ w_{ij} \}$ to fuse the rotated normals.
\end{itemize}

The process is then repeated for $N_\text{iter}$ times. In the following section, we explain how such inference can be done in a convolutional recurrent neural network.


\subsection{Network architecture}
\label{sec:approach-arch}

The components described in Sec.~\ref{sec:approach-ray}-\ref{sec:approach-nrn} are general and can be adopted by most dense prediction neural networks with minimal architectural changes. We use a light-weight CNN with a bottleneck recurrent unit (see Fig.~\ref{fig:method-arch}). The architecture is the same as that of \cite{2022_IronDepth} except for the quantities that are estimated from the updated hidden state.

The initial prediction and the hidden state have the resolution of ($H/8 \times W/8$), where $H$ and $W$ are the input height and width. Updating the normals in a coarse resolution allows us to model long-range relationships with small compute. We set the neighborhood size $\beta$ (mentioned in Sec.~\ref{sec:approach-nrn}) to 2 (i.e. $5 \times 5$ neighborhood). The number of surface normal updates $N_\text{iter}$ is set to 5, as it gave a good balance between accuracy and computational efficiency. As a result, each forward pass returns $N_\text{iter}+1$ predictions (initial prediction obtained via direct regression + $N_\text{iter}$ updates). We then apply convex upsampling~\cite{RAFT} to recover full-resolution outputs (more details regarding the network architecture are provided in the supplementary material). The network is trained by minimizing the weighted sum of their angular losses. The loss for pixel $i$ can be written as

\begin{equation}
\label{eqn:loss}
\mathcal{L}_i = 
\sum_{t=0}^{N_\text{iter}}
\gamma^{N_\text{iter} - t}
\cos^{-1} (\mathbf{n}^\text{gt}_i \cdot \mathbf{n}^t_i )
\end{equation}

\noindent
where $0 < \gamma < 1$ puts a bigger emphasis on the final prediction. We set $\gamma=0.8$ following RAFT~\cite{RAFT}.


\subsection{Dataset}
\label{sec:approach-dataset}

\begin{table}[t]
\footnotesize
\setlength\tabcolsep{1.5pt}
\begin{center}
\begin{tabular}{c|cc|cc}
\hline
\toprule
\multirow{2}{*}{Dataset} & \multicolumn{2}{c|}{Train} & \multicolumn{2}{c}{Val} \\
\cline{2-5}
& \# scenes & \# imgs & \# scenes & \# imgs \\
\midrule
Cleargrasp~\cite{2020_dataset_cleargrasp} & 9 & 900 & 9 & 45\\
3D Ken Burns~\cite{2019_dataset_kenburns} & 23 & 4600 & 23 & 230 \\
Hypersim~\cite{2021_dataset_hypersim} & 407 & 38744 & 407 & 2035 \\
SAIL-VOS 3D~\cite{2021_dataset_sailvos3d} & 170 & 16262 & 170 & 850 \\
TartanAir~\cite{2020_dataset_tartanair} & 16 & 3200 & 16 & 160 \\
MVS-Synth~\cite{2018_dataset_mvssynth} & 120 & 11400 & 120 & 600 \\
BlendedMVG~\cite{2020_dataset_blendedmvs} & 495 & 44070 & 7 & 35 \\
Taskonomy$^\star$~\cite{2018_dataset_taskonomy} & 375 & 37500 & 73 & 365\\
Replica$^\star$~\cite{2019_dataset_replica} & 10 & 1000 & 4 & 20\\
Replica + GSO$^\star$~\cite{2019_dataset_replica,2021_Omnidata} & 30 & 3000 & 12 & 60 \\
\hline
Total & 1655 & 160676 & 841 & 4400 \\
\bottomrule
\end{tabular}
\end{center}
\caption{\textbf{Dataset statistics.} We created a small meta-dataset that covers diverse scenes ($^\star$: downloaded from Omnidata~\cite{2021_Omnidata}).}
\label{table:approach-dataset}
\end{table}

The proposed model is designed to have high sample efficiency. Firstly, we use a fully convolutional design to allow translational weight sharing, which is known to improve the sample efficiency~\cite{2021_Convit}. Secondly, we estimate the rotation matrices by decomposing them into \textit{angles} and \textit{axes}. The angle between two normals is unaffected by the camera pose. While the axis of rotation does change with the camera pose, we estimate the \textit{2D orientation} of its projection on the image plane, which can be as simple as edge detection (its 3D orientation is then recovered from the surface normal). For such reasons, we do not need a large number of images from the same scene. Rendering synthetic scenes with diverse camera intrinsics is also unnecessary as the intrinsics are explicitly encoded in the input.

To this end, we created a small meta-dataset consisting of images extracted from 10 RGB-D datasets (see Tab.~\ref{table:approach-dataset} for dataset composition). Our dataset, compared to Omnidata~\cite{2021_Omnidata}, has a similar number of scenes (1655 vs. 1905) but a significantly smaller number of images (160K vs. 12M).
\section{Experiments}
\label{sec:exp}

After providing details regarding the experimental setup (Sec.~\ref{sec:exp-setup}), we compare the generalization capability of our method to that of the state-of-the-art methods (Sec.~\ref{sec:exp-sota}) and perform an ablation study to demonstrate the effectiveness of the proposed usage of additional inductive biases (Sec.~\ref{sec:exp-ablation}).

\subsection{Experimental setup}
\label{sec:exp-setup}

\textbf{Evaluation protocol.} We measure the angular error for the pixels with ground truth and report the mean and median (lower the better). We also report the percentage of pixels with an error below $t \in [5.0^\circ, 7.5^\circ, 11.25^\circ, 22.5^\circ, 30.0^\circ]$ (higher the better).

\noindent
\textbf{Data preprocessing.} The training images are randomly resized and cropped with a random aspect ratio to facilitate the learning of ray direction-aware prediction. As many of our training images are synthetic, we also add aggressive 2D data augmentation --- e.g., Gaussian blur, Gaussian noise, motion blur, and color --- to minimize the domain gap. Full details regarding data preprocessing are provided in the supplementary material.

\noindent
\textbf{Implementation details.} Our model is implemented in PyTorch~\cite{PyTorch}. In all our experiments, the network and its variants are trained on our meta-dataset (Sec.~\ref{sec:approach-dataset}) for five epochs. We use the AdamW optimizer~\cite{AdamW_introduced} and schedule the learning rate using 1cycle policy~\cite{1cycle-lr} with $lr_\text{max} = 3.5 \times 10^{-4}$. The batch size is set to 4 and the gradients are accumulated every 4 batches. The training approximately takes 12 hours on a single NVIDIA 4090 GPU.

\subsection{Comparison to the state-of-the-art}
\label{sec:exp-sota}

\begin{table*}[t!]
\footnotesize
\setlength\tabcolsep{1.5pt}
\begin{center}
\begin{tabular}{l|cc|ccccc|cc|ccccc|cc|ccccc}
\toprule
\multirow{2}{*}{Method} 
& \multicolumn{7}{c|}{NYUv2~\cite{DATA_2012_NYUv2}}
& \multicolumn{7}{c|}{ScanNet~\cite{DATA_2017_ScanNet}}
& \multicolumn{7}{c}{iBims-1~\cite{DATA_2018_iBims}} \\
\cline{2-22}
& mean & med & {\scriptsize $5.0^{\circ}$} & {\scriptsize $7.5^{\circ}$} & {\scriptsize $11.25^{\circ}$} & {\scriptsize $22.5^{\circ}$} & {\scriptsize $30^{\circ}$} 
& mean & med & {\scriptsize $5.0^{\circ}$} & {\scriptsize $7.5^{\circ}$} & {\scriptsize $11.25^{\circ}$} & {\scriptsize $22.5^{\circ}$} & {\scriptsize $30^{\circ}$} 
& mean & med & {\scriptsize $5.0^{\circ}$} & {\scriptsize $7.5^{\circ}$} & {\scriptsize $11.25^{\circ}$} & {\scriptsize $22.5^{\circ}$} & {\scriptsize $30^{\circ}$} \\
\midrule
OASIS~\cite{2020_dataset_oasis} & 
29.2 & 23.4 & 7.5 & 14.0 & 23.8 & 48.4 & 60.7 & 
32.8 & 28.5 & 3.9 & 8.0 & 15.4 & 38.5 & 52.6 & 
32.6 & 24.6 & 7.6 & 13.8 & 23.5 & 46.6 & 57.4 \\
EESNU~\cite{SNfromRGB_2021_EESNU} & 
{\cellcolor{green!30}}\textbf{16.2} & 8.5 & {\cellcolor{green!30}}\textbf{32.8} & 46.0 & 58.6 & 77.2 & {\cellcolor{green!30}}\textbf{83.5} & 
\textcolor{red}{11.8} & \textcolor{red}{5.7} & \textcolor{red}{45.2} & \textcolor{red}{59.7} & \textcolor{red}{71.3} & \textcolor{red}{85.5} & \textcolor{red}{89.9} & 
20.0 & 8.4 & 32.0 & 46.1 & 58.5 & 73.4 & 78.2 \\
Omnidata v1~\cite{2021_Omnidata} & 
23.1 & 12.9 & 21.6 & 33.4 & 45.8 & 66.3 & 73.6 & 
22.9 & 12.3 & 21.5 & 34.5 & 47.4 & 66.1 & 73.2 & 
19.0 & 7.5 & 37.2 & 50.0 & 62.1 & 76.1 & 80.1 \\
Omnidata v2~\cite{2022_3DCC} & 
17.2 & 9.7 & 25.3 & 40.2 & 55.5 & 76.5 & 83.0 & 
{\cellcolor{green!30}}\textbf{16.2} & 8.5 & 29.1 & 44.9 & 60.2 & {\cellcolor{green!30}}\textbf{79.5} & {\cellcolor{green!30}}\textbf{84.7} & 
18.2 & 7.0 & 38.9 & 52.2 & 63.9 & 77.4 & 81.1 \\
\hline
Ours & 16.4 & 
{\cellcolor{green!30}}\textbf{8.4} & {\cellcolor{green!30}}\textbf{32.8} & {\cellcolor{green!30}}\textbf{46.3} & {\cellcolor{green!30}}\textbf{59.6} & {\cellcolor{green!30}}\textbf{77.7} & {\cellcolor{green!30}}\textbf{83.5} & 
{\cellcolor{green!30}}\textbf{16.2} & {\cellcolor{green!30}}\textbf{8.3} & {\cellcolor{green!30}}\textbf{29.8} & {\cellcolor{green!30}}\textbf{45.9} & {\cellcolor{green!30}}\textbf{61.0} & 78.7 & 84.4 & 
{\cellcolor{green!30}}\textbf{17.1} & {\cellcolor{green!30}}\textbf{6.1} & {\cellcolor{green!30}}\textbf{43.6} & {\cellcolor{green!30}}\textbf{56.5} & {\cellcolor{green!30}}\textbf{67.4} & {\cellcolor{green!30}}\textbf{79.0} & {\cellcolor{green!30}}\textbf{82.3} \\
\midrule
\multirow{2}{*}{\small Method} 
& \multicolumn{7}{c|}{Sintel~\cite{DATA_2012_SINTEL}}
& \multicolumn{7}{c|}{Virtual KITTI~\cite{DATA_2016_VKITTI}}
& \multicolumn{7}{c}{OASIS~\cite{2020_dataset_oasis}} \\
\cline{2-22}
& mean & med & {\scriptsize $5.0^{\circ}$} & {\scriptsize $7.5^{\circ}$} & {\scriptsize $11.25^{\circ}$} & {\scriptsize $22.5^{\circ}$} & {\scriptsize $30^{\circ}$} 
& mean & med & {\scriptsize $5.0^{\circ}$} & {\scriptsize $7.5^{\circ}$} & {\scriptsize $11.25^{\circ}$} & {\scriptsize $22.5^{\circ}$} & {\scriptsize $30^{\circ}$} 
& mean & med & {\scriptsize $5.0^{\circ}$} & {\scriptsize $7.5^{\circ}$} & {\scriptsize $11.25^{\circ}$} & {\scriptsize $22.5^{\circ}$} & {\scriptsize $30^{\circ}$} \\
\midrule
OASIS~\cite{2020_dataset_oasis} & 
43.1 & 39.5 & 1.4 & 3.1 & 7.0 & 24.1 & 35.7 & 
41.8 & 34.6 & 2.7 & 10.1 & 23.6 & 40.8 & 46.7 & 
\textcolor{red}{23.9} & \textcolor{red}{18.2} & \textcolor{red}{-} & \textcolor{red}{-} & \textcolor{red}{31.2} & \textcolor{red}{59.5} & \textcolor{red}{71.8} \\
EESNU~\cite{SNfromRGB_2021_EESNU} & 
42.1 & 36.5 & 3.0 & 6.1 & 11.5 & 29.8 & 41.2 &
51.9 & 53.3 & 1.3 & 4.5 & 14.9 & 29.1 & 34.0 & 
27.7 & 21.0 & - & - & 24.0 & 53.2 & 66.6 \\
Omnidata v1~\cite{2021_Omnidata} & 
41.5 & 35.7 & 3.0 & 5.8 & 11.4 & 30.4 & 42.0 & 
41.2 & 34.0 & 21.5 & 29.3 & 34.7 & 43.0 & 47.6 & 
24.9 & {\cellcolor{green!30}}\textbf{18.0} & - & - & {\cellcolor{green!30}}\textbf{31.0} & 59.5 & 71.4 \\
Omnidata v2~\cite{2022_3DCC} & 
40.5 & 35.1 & 4.6 & 7.9 & 14.7 & 33.0 & 43.5 & 
37.5 & 27.4 & 30.7 & 36.1 & 39.7 & 47.1 & 51.5 & 
{\cellcolor{green!30}}\textbf{24.2} & 18.2 & - & - & 27.7 & {\cellcolor{green!30}}\textbf{61.0} & {\cellcolor{green!30}}\textbf{74.2} \\
\hline
Ours & 
{\cellcolor{green!30}}\textbf{34.9} & {\cellcolor{green!30}}\textbf{28.1} & {\cellcolor{green!30}}\textbf{8.9} & {\cellcolor{green!30}}\textbf{14.1} & {\cellcolor{green!30}}\textbf{21.5} & {\cellcolor{green!30}}\textbf{41.5} & {\cellcolor{green!30}}\textbf{52.7} & 
{\cellcolor{green!30}}\textbf{28.9} & {\cellcolor{green!30}}\textbf{9.9} & {\cellcolor{green!30}}\textbf{43.7} & {\cellcolor{green!30}}\textbf{47.5} & {\cellcolor{green!30}}\textbf{51.3} & {\cellcolor{green!30}}\textbf{59.2} & {\cellcolor{green!30}}\textbf{63.2} & 
24.4 & 18.8 & {\cellcolor{green!30}}\textbf{10.5} & {\cellcolor{green!30}}\textbf{17.5} & 28.8 & 58.5 & 72.0 \\
\bottomrule
\end{tabular}
\end{center}
\caption{\textbf{Quantitative evaluation of the generalization capabilities possessed by different methods.} For each metric, the best results are colored in \textcolor{mygreen}{green}. For evaluation on \cite{DATA_2012_NYUv2, DATA_2017_ScanNet, DATA_2018_iBims, DATA_2012_SINTEL, DATA_2016_VKITTI}, we used the official code and model weights to generate predictions and measured their accuracies. For methods that assume a specific aspect ratio and resolution, the images were zero-padded and resized accordingly to match the requirements. The numbers in \textcolor{myred}{red} mean that the method was trained on the same dataset. We excluded such methods in ranking to ensure a fair comparison.}
\label{table:benchmark}
\end{table*}

\begin{figure*}[t]
\centering
\includegraphics[width=1.0\linewidth]{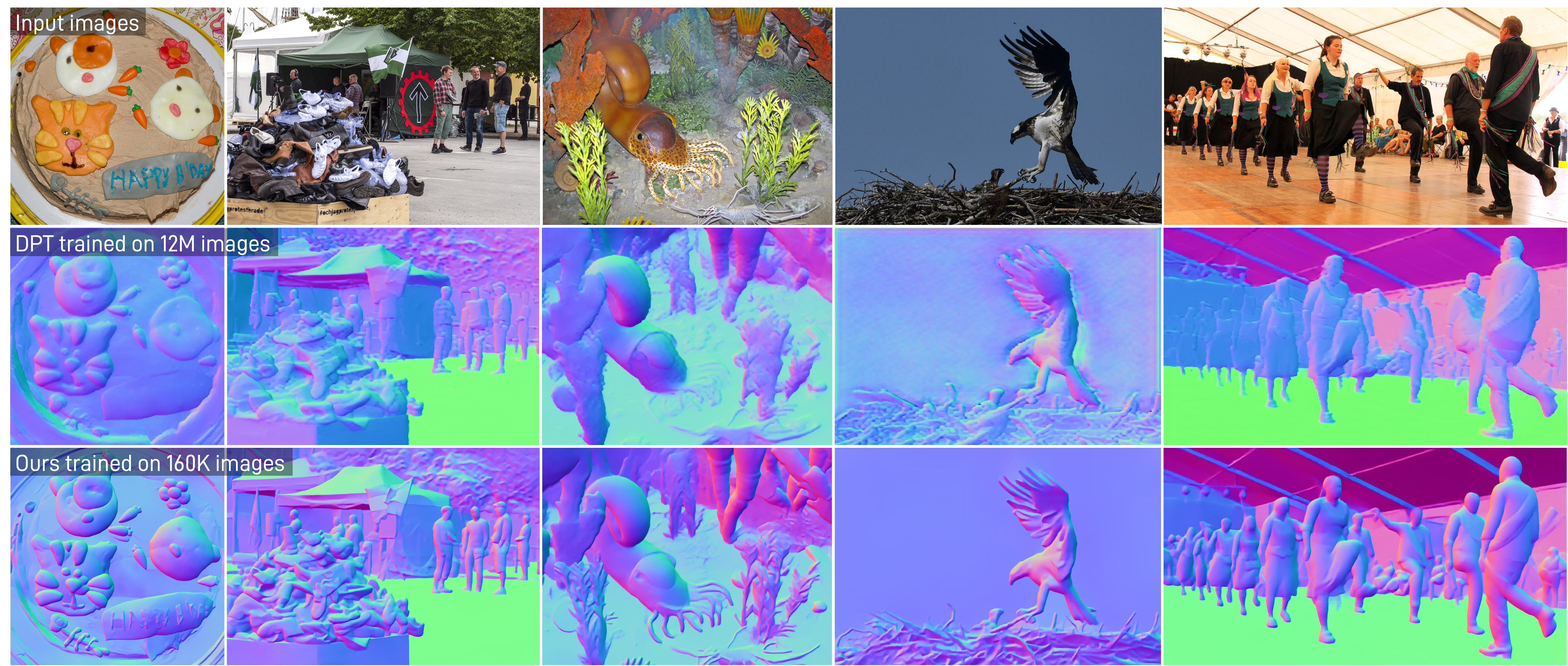}
\caption{\textbf{Comparison to Omnidata v2~\cite{2021_Omnidata} (DPT~\cite{MDE_2021_DPT} model trained on 12 million images using 3D data augmentation~\cite{2022_3DCC} and cross-task consistency~\cite{2020_XTC}).} Our method shows a stronger generalization capability for challenging in-the-wild objects. For texture-less regions (e.g. sky in the fourth column), our model resolves any inconsistency in the prediction and outputs a flat surface, while preserving sharp boundaries around other objects.}
\label{fig:exp-comparison}
\end{figure*}

We select six datasets to compare our method's generalization capability against the state-of-the-art methods. NYUv2~\cite{DATA_2012_NYUv2}, ScanNet~\cite{DATA_2017_ScanNet}, and iBims-1~\cite{DATA_2018_iBims} all contain images of real indoor scenes captured with cameras of standard intrinsics (480 $\times$ 640 resolution and approximately $60^\circ$ field of view). Generalizing to such datasets is straightforward as the scenes and the cameras are similar to those of commonly-used training datasets (e.g. taskonomy~\cite{2018_dataset_taskonomy}). While our method outperforms other methods on most metrics, the improvement is relatively small for this reason.

On the contrary, Sintel~\cite{DATA_2012_SINTEL} and Virtual KITTI~\cite{DATA_2016_VKITTI} contain highly dynamic outdoor scenes and have less common fields of views (e.g. $18^\circ$ to $83^\circ$ for Sintel). The aspect ratios --- $436 \times 1024$ and $375 \times 1242$, respectively --- are also out of distribution. For such datasets, our approach significantly outperforms the other methods across all metrics. This is mainly due to the explicit encoding of the ray direction in the input.

Lastly, we evaluate the methods on the validation set of OASIS~\cite{2020_dataset_oasis}, which contains 10,000 in-the-wild images collected from the internet. Two things should be noted for this dataset. Firstly, the ground truth surface normals only exist for small patches of the images and are annotated by \textit{humans}. Plus, the ground truth is generally available only for large flat regions. The accuracy metrics thus do not faithfully represent the performance of the methods. Secondly, unlike most RGB-D datasets, the camera intrinsics are not available for the input images. We thus approximated the intrinsics by using the focal length recorded in the image metadata, and assuming that the principal point is at the center and that there is zero distortion. Despite such approximation, our method performs on par with the other methods.

For OASIS, we provide a qualitative comparison against Omnidata v2~\cite{2022_3DCC} in Fig.~\ref{fig:exp-comparison}. While the quantitative accuracy was better for \cite{2022_3DCC}, the predictions made by our method show a significantly higher level of detail.

One notable advantage of our method over ViT-based models (e.g. \cite{2022_3DCC}) lies in the simplicity and efficiency of network training. For example, Omnidata v2~\cite{2022_3DCC} was trained for 2 weeks on four NVIDIA V100 GPUs. A set of sophisticated 3D data augmentation functions~\cite{2022_3DCC} were used to improve the generalization performance and cross-task consistency~\cite{2020_XTC} was enforced by utilizing other ground truth labels. On the contrary, our model can be trained in just 12 hours on a single NVIDIA 4090 GPU, does not require geometry-aware 3D augmentations, and does not require any additional supervisory signal. Our model also has 40\% fewer parameters compared to \cite{2022_3DCC} (72M vs 123M). 

\subsection{Ablation study}
\label{sec:exp-ablation}

\begin{table*}[t]
\footnotesize
\setlength\tabcolsep{1.5pt}
\begin{center}
\begin{tabular}{l|c|ccc|c|ccc|c|ccc|c|ccc|c|ccc}
\toprule
\multirow{2}{*}{Method}  
& \multicolumn{4}{c|}{NYUv2~\cite{DATA_2012_NYUv2}}
& \multicolumn{4}{c|}{ScanNet~\cite{DATA_2017_ScanNet}}
& \multicolumn{4}{c|}{iBims-1~\cite{DATA_2018_iBims}} 
& \multicolumn{4}{c|}{Sintel~\cite{DATA_2012_SINTEL}}
& \multicolumn{4}{c}{Virtual KITTI~\cite{DATA_2016_VKITTI}} \\
\cline{2-21}
& mean & {\scriptsize $11.25^{\circ}$} & {\scriptsize $22.5^{\circ}$} & {\scriptsize $30^{\circ}$} 
& mean & {\scriptsize $11.25^{\circ}$} & {\scriptsize $22.5^{\circ}$} & {\scriptsize $30^{\circ}$} 
& mean & {\scriptsize $11.25^{\circ}$} & {\scriptsize $22.5^{\circ}$} & {\scriptsize $30^{\circ}$}
& mean & {\scriptsize $11.25^{\circ}$} & {\scriptsize $22.5^{\circ}$} & {\scriptsize $30^{\circ}$}
& mean & {\scriptsize $11.25^{\circ}$} & {\scriptsize $22.5^{\circ}$} & {\scriptsize $30^{\circ}$} \\
\midrule
baseline & 16.6 & 59.1 & 76.8 & 82.9 & 16.5 & 60.5 & 77.7 & 83.5 & 18.0 & 66.0 & 77.7 & 81.2 & 36.6 & 18.4 & 38.3 & 50.0 & 30.5 & 48.0 & 55.9 & 60.6 \\
baseline + ray & 16.6 & 59.2 & 77.0 & 82.9 & 16.4 & 61.2 & 77.9 & 83.6 & 17.6 & 66.4 & 78.1 & 81.4 & 36.0 & 19.8 & 38.8 & 50.4 & 29.4 & 50.6 & 58.4 & 62.4 \\
baseline + ray + rot & 
{\cellcolor{green!30}}\textbf{16.4} & {\cellcolor{green!30}}\textbf{59.6} & {\cellcolor{green!30}}\textbf{77.7} & {\cellcolor{green!30}}\textbf{83.5} & 
{\cellcolor{green!30}}\textbf{16.2} & {\cellcolor{green!30}}\textbf{61.0} & {\cellcolor{green!30}}\textbf{78.7} & {\cellcolor{green!30}}\textbf{84.4} & {\cellcolor{green!30}}\textbf{17.1} & {\cellcolor{green!30}}\textbf{67.4} & {\cellcolor{green!30}}\textbf{79.0} & {\cellcolor{green!30}}\textbf{82.3} & {\cellcolor{green!30}}\textbf{34.9} & {\cellcolor{green!30}}\textbf{21.5} & {\cellcolor{green!30}}\textbf{41.5} & {\cellcolor{green!30}}\textbf{52.7} & {\cellcolor{green!30}}\textbf{28.9} & {\cellcolor{green!30}}\textbf{51.3} & {\cellcolor{green!30}}\textbf{59.2} & {\cellcolor{green!30}}\textbf{63.2} \\
\bottomrule
\end{tabular}
\end{center}
\caption{\textbf{Ablation study - quantitative results.} Adding per-pixel ray direction as input (+ ray) and updating the initial prediction via iterative rotation estimation (+ rot) both lead to an overall improvement in the metrics. The benefit of using ray direction encoding is clearer for datasets with out-of-distribution camera intrinsics (Sintel and Virtual KITTI).}
\label{table:ablation}
\end{table*}

\begin{figure}[t]
\centering
\includegraphics[width=1.0\linewidth]{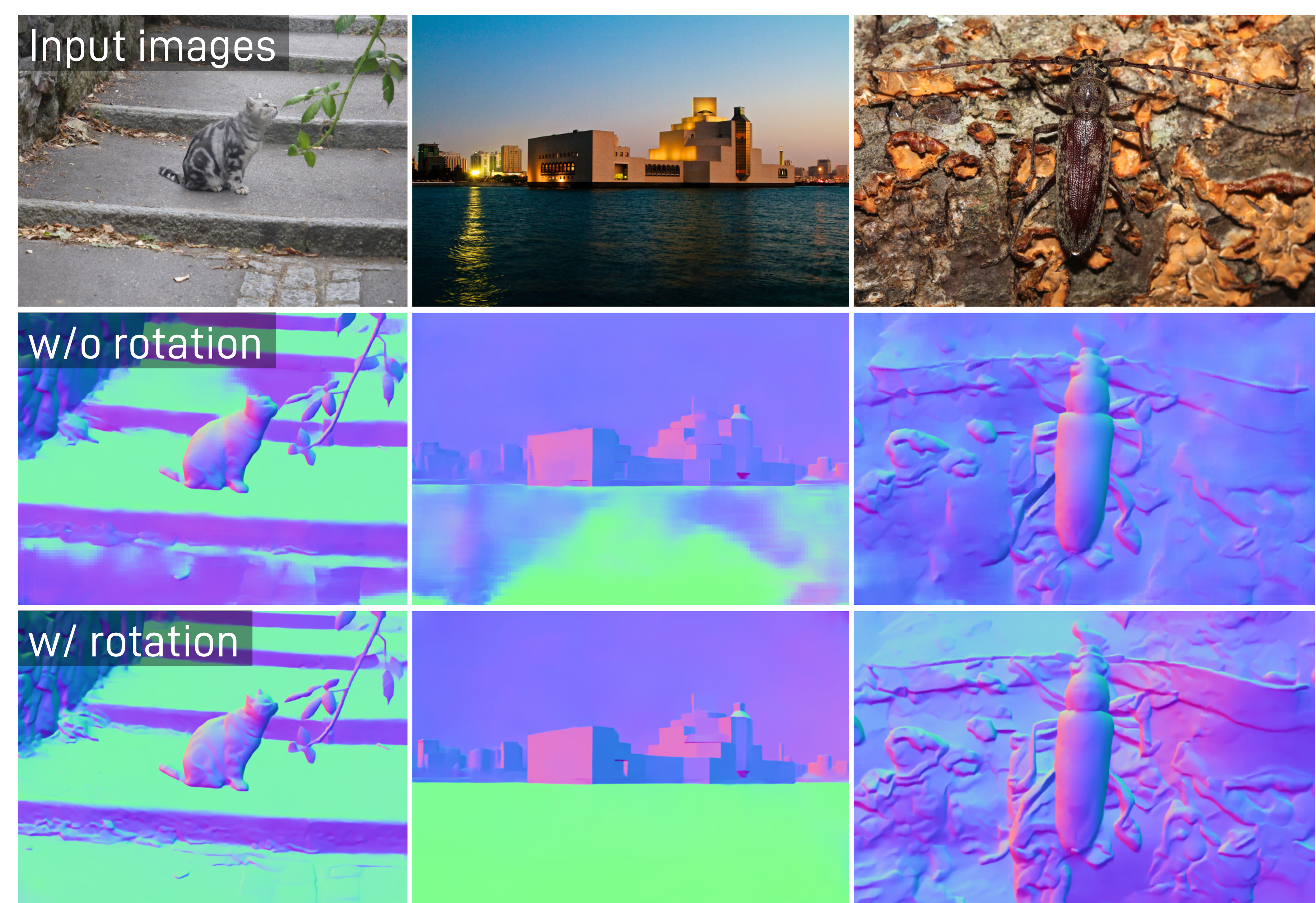}
\caption{\textbf{Ablation study - qualitative results.} Here we compare the predictions made by two models with and without the iterative rotation estimation (both are using per-pixel ray information). When the model is trained to directly estimate the normals, the prediction is often inconsistent within smooth surfaces, leading to \textit{bleeding} artifacts. The proposed refinement via rotation estimation leads to piece-wise smooth surfaces that are crisp near surface boundaries.}
\label{fig:exp-ablation}
\end{figure}

We now run an ablation study to examine the effectiveness of the proposed usage of two new inductive biases --- utilizing dense per-pixel ray direction and modeling the pairwise relative rotation between nearby pixels. As can be seen from Tab.~\ref{table:ablation}, encoding the ray direction helps improve the accuracy, especially for out-of-distribution camera intrinsics (Sintel and Virtual KITTI). This is in line with our observations in Tab.~\ref{table:benchmark}.

On the other hand, the improvement coming from the rotation estimation is not big. As the accuracy metrics are dominated by the pixels belonging to large planar surfaces, they do not convey the improvements near surface boundaries. The metrics also do not penalize the inconsistencies within piece-wise smooth surfaces. Qualitative comparison in Fig.~\ref{fig:exp-ablation} clearly shows that the proposed refinement via rotation estimation improves the piece-wise consistency and the sharpness of the prediction near surface boundaries.
\section{Conclusion}
\label{sec:conclusion}

In this paper, we discussed the inductive biases needed for surface normal estimation and introduced how per-pixel ray direction and the relative rotational relationship between neighboring pixels can be encoded in the output. Per-pixel ray direction allows camera intrinsics-aware inference and thus improves the generalization ability, especially when tested on images taken with out-of-distribution cameras. Explicit modeling of inter-pixel constraints --- implemented in the form of rotation estimation --- leads to piece-wise smooth predictions that are crisp near object boundaries. 

Compared to a recent transformer-based state-of-the-art method, our method shows stronger generalization capability and a significantly higher level of detail in the prediction, despite being trained on an orders of magnitude smaller dataset. Thanks to its fully convolutional architecture, our model can be applied to images of arbitrary resolution and aspect ratio, without the need for image resizing or position encoding inter/extrapolation. We believe that the domain- and camera-agnostic generalization capability of our method makes it a strong front-end perception that can benefit many downstream 3D computer vision tasks.
\section{Limitation and future work}
\label{sec:future}

Surface normal estimation is an inherently ambiguous task when the camera intrinsics are not known. This was why we proposed to encode the camera intrinsics in the form of dense per-pixel ray direction. While this helped us push the limits of single-image surface normal estimation, it also means that the model requires prior knowledge about the camera. 

Note, however, that most RGB-D datasets already provide pre-calibrated camera parameters, and that monocular cameras can be calibrated easily using patterns with known relative coordinates. For in-the-wild images, we demonstrated in Sec.~\ref{sec:exp-sota} that the intrinsics can be approximated using the image metadata. If no information is available, we can attempt to \textit{estimate} the camera intrinsics from a single image. For instance, vanishing points with known relative angles can be used to recover the camera parameters~\cite{1990_camera_calibration}. As our model is designed to learn the relative angle between surfaces, it can in turn be used for camera calibration. This will be explored in our future work.
\section{Acknowledgement}
\label{sec:acknowledgement}

Research presented in this paper was supported by Dyson Technology Ltd. The authors would like to thank Shikun Liu, Eric Dexheimer, Callum Rhodes, Aalok Patwardhan, Riku Murai, Hidenobu Matsuki, and members of the Dyson Robotics Lab for insightful feedback and discussions.
{
    \small
    \bibliographystyle{ieeenat_fullname}
    \bibliography{main}
}

\clearpage
\setcounter{page}{1}
\maketitlesupplementary

\section{Network architecture}
\label{sec:supp-arch}

\begin{table*}[h]
\setlength{\tabcolsep}{2.0pt}
\begin{center}
\begin{tabular}{c|c|c|c}
\toprule
\textbf{Input} & \textbf{Layer} & \textbf{Output} & \textbf{Output Dimension} \\
\hline
\textit{image} & - & - & $H \times W \times 3$ \\
\hline
\multicolumn{4}{c}{\textbf{Encoder}} \\
\hline
\multirow{3}{*}{\textit{image}}
& \multirow{3}{*}{EfficientNet B5}
& \textit{$F_8$} & $H/8 \times W/8 \times 64$ \\
& & \textit{$F_{16}$} & $H/16 \times W/16 \times 176$ \\
& & \textit{$F_{32}$} & $H/32 \times W/32 \times 2048$ \\
\hline
\multicolumn{4}{c}{\textbf{Decoder}} \\
\hline
\textit{$F_{32}$} + $\mathbf{r}_{32}$ & Conv2D(ks=1, $C_\text{out}$=2048, padding=0) 
& $x_0$
& $H/32 \times W/32 \times 2048$ \\
\hline
$\text{up}(x_0) + F_{16} + \mathbf{r}_{16}$
& 
$\left( 
\begin{matrix} 
\text{Conv2D(ks=3, $C_\text{out}$=1024, padding=1)}, \\
\text{GroupNorm}(n_\text{groups}=8), \\
\text{LeakyReLU()}
\end{matrix}
\right)
\times 2$ 
& $x_1$ & $H/16 \times W/16 \times 1024$ \\
\hline
$\text{up}(x_1) + F_{8} + \mathbf{r}_{8}$
& 
$\left( 
\begin{matrix} 
\text{Conv2D(ks=3, $C_\text{out}$=512, padding=1)}, \\
\text{GroupNorm}(n_\text{groups}=8), \\
\text{LeakyReLU()}
\end{matrix}
\right)
\times 2$ 
& $x_2$ & $H/8 \times W/8 \times 512$ \\
\hline
\multicolumn{4}{c}{\textbf{Prediction Heads}} \\
\hline
$x_2 + \mathbf{r}_{8}$ & 
\makecell{
Conv2D(ks=3, $C_\text{out}$=128, padding=1), ReLU(), \\
Conv2D(ks=1, $C_\text{out}$=128, padding=0), ReLU(), \\
Conv2D(ks=1, $C_\text{out}$=3, padding=0), Normalize(), viewReLU() \\
}
& $\mathbf{n}^{t=0}$ & $H/8 \times W/8 \times 3$ \\
\hline
$x_2 + \mathbf{r}_{8}$ & 
\makecell{
Conv2D(ks=3, $C_\text{out}$=128, padding=1), ReLU(), \\
Conv2D(ks=1, $C_\text{out}$=128, padding=0), ReLU(), \\
Conv2D(ks=1, $C_\text{out}$=64, padding=0) \\
}
& $\mathbf{f}$ & $H/8 \times W/8 \times 64$ \\
\hline
$x_2 + \mathbf{r}_{8}$ & 
\makecell{
Conv2D(ks=3, $C_\text{out}$=128, padding=1), ReLU(), \\
Conv2D(ks=1, $C_\text{out}$=128, padding=0), ReLU(), \\
Conv2D(ks=1, $C_\text{out}$=64, padding=0) \\
}
& $\mathbf{h}^{t=0}$ & $H/8 \times W/8 \times 64$ \\
\bottomrule
\end{tabular}
\end{center}
\caption{\textbf{Network architecture.} In each 2D convolutional layer, "ks" and $C_\text{out}$ are the kernel size and the number of output channels, respectively. $F_N$ represents the feature-map of resolution $H/N \times W/N$, and $\mathbf{r}_N$ is a dense map of per-pixel ray direction in the same resolution. $X + Y$ means that the two tensors are concatenated, and $\text{up}(\cdot)$ is bilinear upsampling by a factor of 2.}
\label{table:supp-arch}
\end{table*}

Tab.~\ref{table:supp-arch} shows the architecture of the CNN used to extract the initial surface normals, initial hidden state, and context feature. For the ConvGRU cell and convex upsampling layer, we use the architecture of \cite{2022_IronDepth} and \cite{RAFT}, respectively.

\section{Data preprocessing}
\label{sec:supp-data}

During training, the input image goes through the following set of data augmentation ($p$: the probability of applying each augmentation).

\begin{itemize}

\item \textbf{Downsample-and-upsample} ($p=0.1$). Bilinearly downsample the image $(H \times W)$ into $(rH \times rW)$, where $r \sim \mathcal{U}(0.2, 1.0)$. Then upsample it back to $(H \times W)$.
    
\item \textbf{JPEG compression} ($p=0.1$). Apply JPEG compression with quality $q \sim \mathcal{U}(10, 90)$.

\item \textbf{Gaussian blur} ($p=0.1$). Add Gaussian blur with kernel size $(11 \times 11)$ and $\sigma \sim \mathcal{U}(0.1, 5.0)$.

\item \textbf{Motion blur} ($p=0.1$). Simulate motion blur by convolving the image with a 2D kernel whose value is 1.0 along a line that passes through the center and is 0.0 elsewhere. The kernel is then normalized such that its sum equals 1.0. The kernel size is drawn randomly from $[1, 3, 5, 7, 9, 11]$.

\item \textbf{Gaussian noise} ($p=0.1$). Add Gaussian noise $x \sim \mathcal{N}(0, \sigma)$ where $\sigma \sim \mathcal{U}(0.01, 0.05)$. Note that the image is pre-normalized to $[0.0, 1.0]$.

\item \textbf{Color} ($p=0.1$). Use ColorJitter in PyTorch~\cite{PyTorch} with (brightness=0.5, contrast=0.5, saturation=0.5, hue=0.2).

\item \textbf{Grayscale} ($p=0.01$). Change the image into grayscale.

\end{itemize}

We also randomize the aspect ratio of the input image. Suppose that the input has a resolution of $H \times W$. We first randomize the target aspect ratio $H^\text{target} \times W^\text{target}$, while making sure that the total number of pixels is roughly 300K (to maintain GPU memory usage). We then resize the input into $rH \times rW$, such that $rH \sim \mathcal{U}(\min(H, H^\text{target}), \max(H, H^\text{target}))$. The resized input is then cropped based on the target resolution.

\section{Additional figures and video}
\label{sec:supp-add}

\begin{figure*}[t]
\centering
\includegraphics[width=0.95\linewidth]{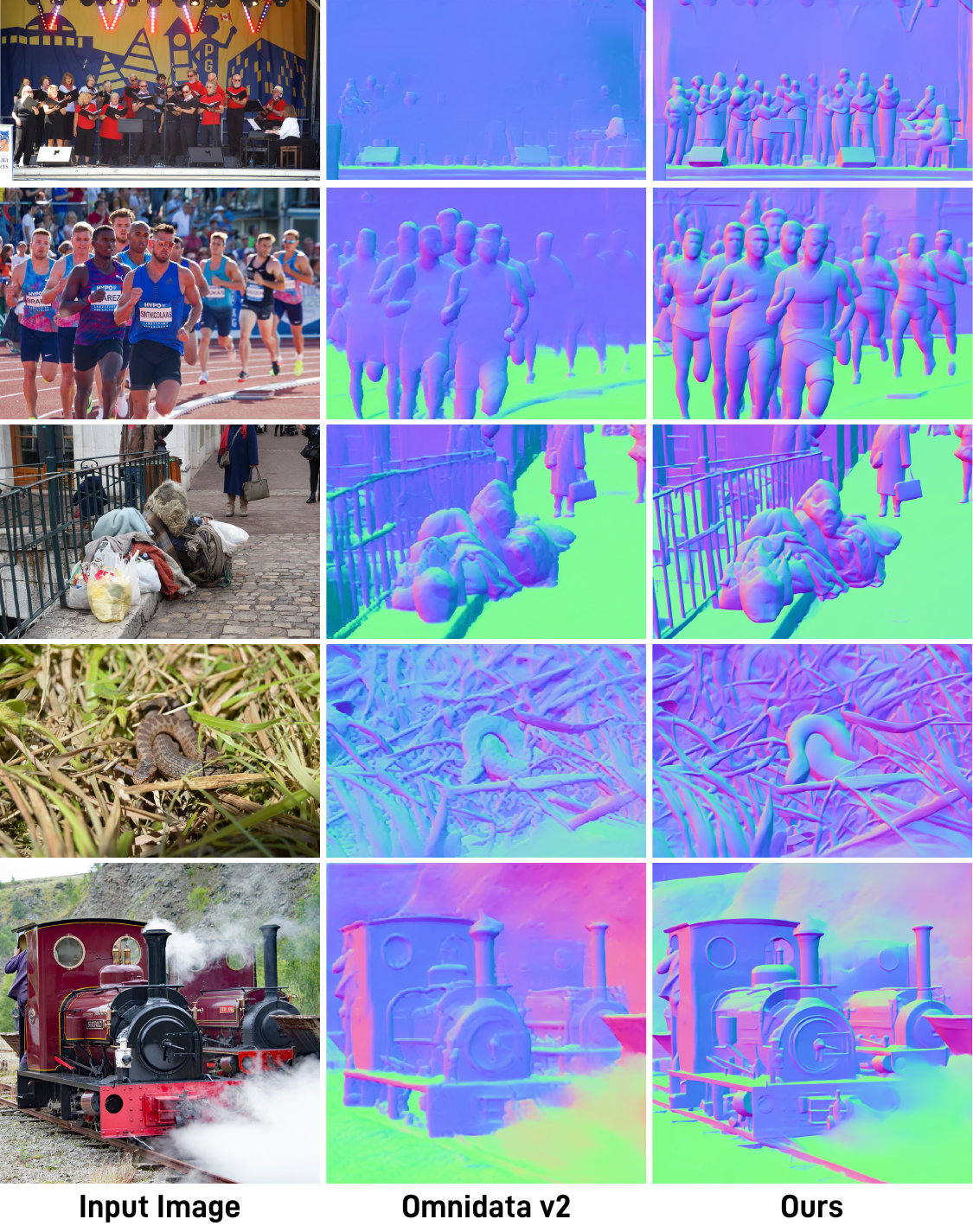}
\caption{\textbf{Additional comparison to Omnidata v2~\cite{2021_Omnidata} on in-the-wild images from the OASIS dataset~\cite{2020_dataset_oasis}.}}
\label{fig:supp-comparison}
\end{figure*}

We provide an additional qualitative comparison to Omnidata v2~\cite{2021_Omnidata} in Fig.~\ref{fig:supp-comparison}. 


\end{document}